\documentclass[runningheads]{llncs}
\usepackage{graphicx}
\usepackage{amsmath,amssymb} % define this before the line numbering.
\usepackage{color}
\usepackage{gensymb}
\usepackage{enumerate}
\usepackage{subfig}
\usepackage[font={small}]{caption}
\usepackage[width=122mm,left=12mm,paperwidth=146mm,height=193mm,top=12mm,paperheight=217mm]{geometry}

\def\ie{\emph{i.e.~}}
\def\eg{\emph{e.g.~}}
\def\etal{\emph{et al.~}}
\def\wrt{\emph{w.r.t.~}}

\newcommand{\mytilde}{\raise.17ex\hbox{$\scriptstyle\mathtt{\sim}$}}

\begin{document}
% \linenumbers

\pagestyle{headings}
\mainmatter
\def\ECCV14SubNumber{638}  % Insert your submission number here

\title{Direction matters: hand pose estimation\\ from local surface normals} % Replace with your title

% \titlerunning{ECCV-16 submission ID \ECCV14SubNumber}

% \authorrunning{ECCV-16 submission ID \ECCV14SubNumber}
\author{Chengde Wan$^{1}$, Angela Yao$^{2}$, Luc Van Gool$^{1,3}$}
\institute{
    $^{1}$  Computer Vision Laboratory, D-ITET, ETH Zurich\\
    $^{2}$  Department of Computer Science, University of Bonn\\
    $^{3}$  VISICS, ESAT, K.U. Leuven\\
    {\tt\small \{wanc,vangool\}@vision.ee.ethz.ch,yao@informatik.uni-bonn.de}
}

% \author{Chengde Wan\inst{1},~
%         Angela Yao\inst{2},~
%         \and Luc Van Gool\inst{1,3}
% }
% \institute{ 
%     \inst{1}~
%     Computer Vision Laboratory, D-ITET, ETH Zurich\\
%     \inst{2}~
%     Department of Computer Science, University of Bonn\\
%     \inst{3}~
%     VISICS, ESAT, K.U. Leuven\\
% }

\maketitle

\begin{abstract}
We present a hierarchical regression framework for estimating hand joint positions from single depth images based on local surface normals. The hierarchical regression follows the tree structured topology of hand from wrist to finger tips. We propose a conditional regression forest, \ie the \emph{Frame Conditioned Regression Forest} (FCRF) which uses a new  normal difference feature. 
At each stage of the regression, the frame of reference is established from either the local surface normal or previously estimated hand joints. By making the regression with respect to the local frame, the pose estimation is more robust to rigid transformations. We also introduce a new efficient approximation to estimate surface normals. We verify the effectiveness of our method by conducting experiments on two challenging real-world  datasets and show consistent improvements over previous discriminative pose estimation methods.
\end{abstract}

\section{Introduction}
We consider the problem of 3D hand pose estimation from single depth images.  Hand pose estimation has important applications in human-computer interaction (HCI) and augmented reality (AR). Estimating the freely moving hand has several challenges including large viewpoint variance, finger similarity and self occlusion and versatile and rapid finger articulation. 
%especially in pitch and roll angles \AY{it's not clear which these might be, doesn't this depend on your reference point?}; 

Methods for hand pose estimation from depth generally fall into two camps.  The first is frame-to-frame model based tracking~\cite{forth-pso,forth-quasi,msra_modelbased,Liang-tip}.  Model-based tracking approaches can be highly accurate if given enough computational resources for the optimization.  The second camp, where our work also falls, is single frame discriminative pose estimation~\cite{cascaded-hand,Tang-LRF,Li2015-sip,xuHand_13}. These methods are less accurate than model-based trackers but much faster and are targeted towards real-time performance without GPUs. Model-based tracking and discriminative pose estimation are complementary to each other and there have been notable hybrid methods  ~\cite{Tang2015-blackbox,msr-chi,detection_guided,collatorative,Ballan-hand} which try to maintain the advantages of both camps.

% \AY{conclude the introduction section, by stating how our work complementse model-based tracking}
% Finally, there have been some notable combination frameworks of late ~\cite{Tang2015-blackbox,msr-chi,detection_guided,collatorative} which try to maintain the advantages from both camps.
% model-based tracking suffer from local minima, our work is complementary, and can also be combined into such a framework.

%As we are interested in the use case of real-time systems, 
%\AY{discuss why directly estimating everything is a bad idea}

Earlier methods for discriminative hand pose estimation tried to estimate all joints directly~\cite{Keskin2012-ns,forth-oneshot} though such approaches tend to fail with dramatic view-point changes and extreme articulations.  Following the lead of several notable methods~\cite{cascaded-hand,Tang-LRF,Li2015-sip,Tang2015-blackbox}, 
we cast pose estimation as a hierarchical regression problem.
%we regress the hand joints hierarchically.  % from the wrist to the finger tips based on a given hand skeleton model. 
The idea is to start with easier parent parts such as the wrist or palm, and then tackle subsequent and more difficult children parts such as the fingers. The assumption is that the children parts, once conditioned on the parents, will exhibit less variance and simplify the learning task.  % given its parent part and makes the training and testing easier. 
Furthermore, by constraining the underlying graphical model to follow the tree-structured topology of the hand, hierarchical regression implicitly captures the skeleton constraints and therefore shares some advantages of model-based tracking that are otherwise not present when directly estimating all joints independently.
% stepwise constraints / local estimation; implicitly captures some of the constraints and therefore advantages of model-based tracking.
%we cast hand pose estimation as a hierarchical regression problem by regressing the hand joint from wrist to finger tips based on a given model. 
%Fundamental assumptions of (1) independence in each of the finger joints and (2) vote for all possible joints from the evidence
%hierarchical hand pose estimation starts with easier and or more discriminative joints and constrains the estimation subsequently.  This has given 
%\AY{performance gains wrt to previous works}

%We propose a method of ... \AY{one to two paragraphs describing how your method works}
%In this paper, 
%We propose a hierarchical regression framework to estimate hand joint positions from single depth images. We 
Our framework starts with estimating the surface normals of given point clouds. The normal direction establishes the local reference frames used in later conditional regression and serves as features.%a local feature map.  
We then apply our %conditional regression scheme, \ie 
\emph{Frame Conditioned Regression Forest} (FCRF) to hierarchically regress hand joints down from the wrist to the finger tips. At each stage, the frame of reference is established based on previously estimated local surface normal or joint positions. The regression forest considers offsets between input points and joints of interest with respect to the local reference frame and also conditions the feature with respect to these local frames.  Our use of conditioned features is inspired by~\cite{cascaded-hand}, though we consider angular differences between local surface normals, which is far more robust to rigid transformations than the original depth difference feature. 

\begin{figure}[t]
    \centering
    \includegraphics[width=.9\textwidth]{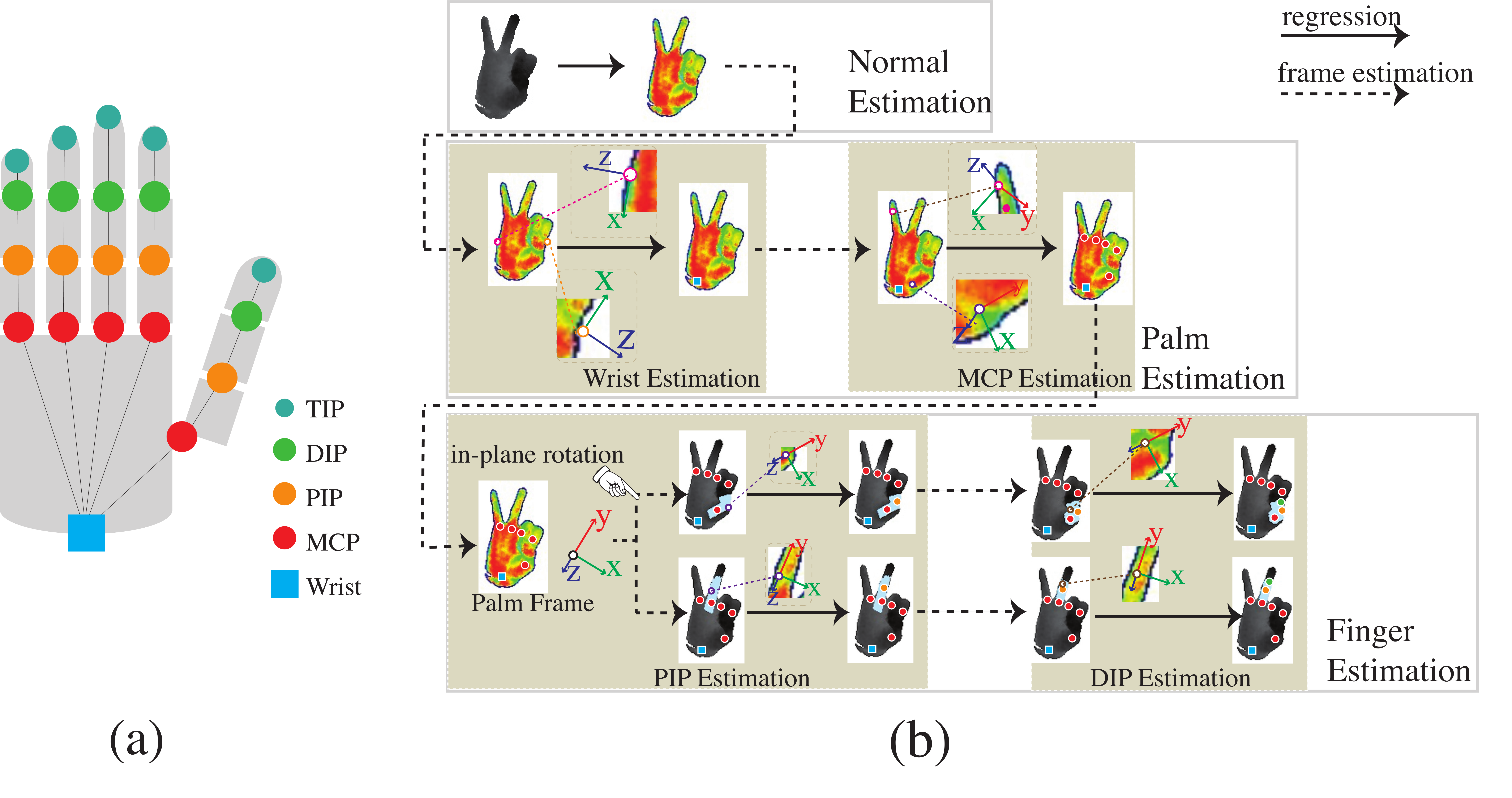}
    \caption{Framework. (a) shows the hand skeleton model used in our work. (b) sketches our hierarchical regression framework, with each successive stage denoted by a shaded box.% based on the tree-structured skeleton in (a). 
    %~\AY{what is dif. btw brown vs. gray?  can we removegray boxes?}
    We first estimate a reference frame for every input point encoding all  information from previous stages and use that reference frame as input to estimate the location of children joints.  The sub-figure around the depth map amplifies a local region from the initial depth map and shows the corresponding frame for a specific point. To save space, only thumb and index finger cases are shown and finger tip points(TIP) estimation is omitted as it is identical to that of DIP \textbf{(best viewed in colour)} }
    \label{fig:framework}
    \vspace{-0.5cm}
\end{figure}

Our proposed method has the following contributions:
\begin{enumerate}
    \item We are the first to incorporate local surface normals for pose estimation.  To this end, we propose an extremely efficient normal estimation method based on regression trees adapted to handle unit vector distributions, different from vector space properties.     
%    The local surface normal helps to establish the local reference frame.   
    \item We extend the commonly used depth difference feature\cite{cascaded-hand,Tang-LRF,Li2015-sip,Tang2015-blackbox,Shotton2013_pami,Vitruvian} to an angular difference feature between two normal directions.  Our normal difference feature is highly robust to 3D rigid transformation.  In particular, the feature is invariant to in-plane rotations, which means we can dispense with data augmentation and have more efficient training and testing routines.  
    \item We propose a flexible conditional regression framework, encoding all previously estimated information as a part of the local reference frame.  This includes local point properties such as the normal direction and global properties such as the estimated joint position.
\end{enumerate}

%\AY{IMO, outline of the rest of the paper is not necessary; depends on personal taste.  Instead, comment a bit about how well method compares wrt to state of the art approaches.}
We validate our method on two real-world challenging hand pose estimation datasets, ICLV\cite{Tang-LRF} and MSRA\cite{cascaded-hand}. On ICLV, we achieve the state-of-art performance against all previous discriminative based methods~\cite{cascaded-hand,Tang-LRF,Li2015-sip} with a large margin. On MSRA, our method is on-par with the state-of-art methods~\cite{collatorative,cascaded-hand} at the threshold of 40mm, and with some minor modifications outperforms ~\cite{collatorative,cascaded-hand}.

\section{Related Works}
\label{sec:relate}
We limit our discussion to the most relevant issues and works, and refer readers to~\cite{review15,review07} for more comprehensive reviews on hand pose estimation in general. \vspace{-0.3cm}
\paragraph{Hierarchical Regression}
Several methods have adopted some form of hierarchical treatment of the pose estimation problem.  For example, in~\cite{msr-chi,Keskin2012-ns,Tang-sf}, the hand is first classified into several classes according to posture or viewpoint; further pose  estimation is then conditioned on such initial class.  Obviously, such an approach cannot generalize to unseen postures and viewpoints.

Other works~\cite{cascaded-hand,Tang-LRF,Li2015-sip,xuHand_13,Tang2015-blackbox} hierarchically follow the tree-structured hand topology. In~\cite{Tang-LRF,Li2015-sip}, data points are recursively partitioned into subsets and only corresponding subsets of points are considered for subsequent joint estimation.  In~\cite{Tang2015-blackbox}, estimated parent joints are used as inputs for regressing children joints; a final energy minimization is applied to refine the estimation.  In~\cite{cascaded-hand,xuHand_13}, predictions are made based on previously estimated reference frames. Our work is similar in spirit to~\cite{cascaded-hand,xuHand_13}, as we also make estimations based on reference frames.  However, unlike \cite{cascaded-hand,xuHand_13}, we utilize the normal direction to establish the reference frame and take  local point properties into consideration.  Further explanations on the differences between our work and \cite{cascaded-hand,xuHand_13} are given in Section~\ref{sec:ndf} and~\ref{sec:rf}.

\paragraph{Viewpoint Handling} %Unlike in body pose estimation, which is limited to mostly frontal views, 
The free moving hand can exhibit large viewpoint changes and a variety of techniques have been proposed to handle these.  For example,~\cite{Tang-sf,dantone2012real} discretize viewpoints into multiple classes and estimate pose in the view-specific classes. Unfortunately, these methods may introduce quantization errors and cannot generalize to unseen viewpoints. In~\cite{xuHand_13}, the regression for hand pose is conditioned on an estimated in-plane rotation angle.  This is extended in~\cite{cascaded-hand},  
%, which conditions the regression on an estimated 3D pose. More specifically,~\cite{cascaded-hand}, 
which regresses the pose residual iteratively, conditioned on the estimated 3D pose at each iteration. Such a method is highly sensitive to the pose initialization and may get trapped in local minima.  

%starting from an initialization which would make the result trapped to local minimals due to poor initializations.

%We too are conditioning our regression, though 
%Ours is similar to \cite{cascaded-hand} in this aspect.  But ours exhibits two major differences to \cite{cascaded-hand}, 1) we utilize the local surface normal direction to establish the Darboux local frame, 
% while %; 2) instead of depth difference feature, we use the angular difference between to normal vectors, which is highly robust to rigid body transformation and better exploits surface local property.

\paragraph{Point Cloud Features}%Properties}
Depth difference features are widely used together with random forests in body pose~\cite{Shotton2013_pami,Vitruvian} and hand pose~\cite{cascaded-hand,Tang-LRF,Li2015-sip,xuHand_13,Tang2015-blackbox,Keskin2012-ns,Tang-sf} estimation. %The feature is extremely efficient to calculate and its large feature dimensionality pair well with the random forest feature selection power. 
Depth differences, however, ignore many local geometric properties of the point cloud, \eg local surface normals and curvatures, and are not robust to rigid transformations and sensor noise. %, making the feature less discriminative and not robust to rigid transformation and sensor noise.

In ~\cite{msra_modelbased,Liang-tip} geodesic extreme points such as finger tip candidates are used to guide later estimation. 
Rusu \etal~\cite{Rusu-pfh} proposed a histogram feature describing different local properties. Inspired by~\cite{Rusu-pfh}, we establish local Darboux frames and using angular differences as feature values, but unlike~\cite{Rusu-pfh}, our features are based on random offsets and retain the efficiency of~\cite{Shotton2013_pami}. Most recently, convolutional neural networks (CNNs) have been used to automatically learn point cloud features~\cite{graz-cnn,nyu_hand}.  Due to the heavy computational burden, CNNs can still not be used in real-time without a GPU.  %In our work, random forest is firstly used to estimate the local surface normal which is used as feature map later. To this end, our normal estimation random forest is similar to the first several convolutional layers in CNN, which non-linearly regress a new feature map. But ours has explicit geometrical meaning and is used to establish Darboux frame, and is far more efficient then that of CNN.

%   \noindent\textbf{Point local property} Knowing the local property for a certain point from the point cloud is critical in estimating the hand joint position. \cite{nyu_hand} classifies points into non-hand and hand point to exclude noise from background. \cite{msra_modelbased,Liang-tip} detect the geodesic extreme point as the tip point candidate to guide later estimation. \cite{Keskin2012-ns,Tang-sf,xuHand_13} classify every pixel into a hand-class part independently and \cite{Liang-mrf} takes part correlation into account. The position of the joints are estimated based on the hand part classification result. 
%Position of the corresponding joints are estimated only based on a sub-set of points and release the problem of self-similarity. Unlike all previous work for pose estimation, we are the first to use normal direction as the local property. The benefits of the local surface normal direction are two-folds, 1) pose/direction aware, helps to establish local frame; 2) can be considered as feature map, that encodes the local property and the feature difference between two point can be calculated;

\section{Random Normal Difference Feature}
\label{sec:normal}
\subsection{Random difference features}
One of the most commonly used features in depth-based pose estimation frameworks, for both body pose estimation~\cite{Shotton2013_pami,Vitruvian} and hand pose estimation~\cite{cascaded-hand,xuHand_13}, is the random depth difference feature~\cite{Shotton2013_pami}.  Formally, the random difference feature $f_{\mathcal{I}}$ for point $\mathbf{p}_i \in \mathcal{R}^3$ from depth map $\mathcal{I}$ is defined as follows,
\begin{equation}
\label{equ:rdf}
f_{\mathcal{I}}(\mathbf{p}_i, \mathbf{\delta}_1, \mathbf{\delta}_2) = \Delta(\phi_{\mathcal{I}}(r(\mathbf{p}_i, \mathbf{\delta}_1)), \phi_{\mathcal{I}}(r(\mathbf{p}_i, \mathbf{\delta}_2))),
\end{equation}
\noindent
where $\mathbf{\delta}_j \in \mathcal{R}^3, j=\{1,2\}$ is a random offset, $r(\mathbf{p}_i, \mathbf{\delta}_j) \in \mathcal{R}^3$ calculates a random position given point $\mathbf{p}_i$ and offset $\mathbf{\delta}_j$.  $\phi_{\mathcal{I}}(\mathbf{q})$ is the local feature map for position $\mathbf{q}\in \mathcal{R}^3$ on the point cloud and   $\Delta(\cdot, \cdot)$ returns the local feature difference.  In the case of random depth difference features~\cite{cascaded-hand,xuHand_13,Shotton2013_pami}, $\phi_{I}$ is the recorded depth, though the same formalism applies for other features.

Random difference features are well suited for random forest frameworks; the many possible combinations of offsets perfectly utilize their feature selection and generalization power. In addition, every dimension of the feature is calculated independently, which gives rise to parallelization schemes and allows for both temporal and spatial efficiency in training and testing.  One of the main drawbacks of the depth-difference feature, however, is its inability to cope with transformations.  Since random offsets in $r(\mathbf{p}_i, \mathbf{\delta}_1)$ are determined either \textit{w.r.t.} the camera frame~\cite{Shotton2013_pami} or to a globally estimated frame~\cite{cascaded-hand,xuHand_13}, the depth difference for the same offset can vary widely under out of plane rotations.  %, especially to the highly articulated object like hand.

%For example, the depth difference between two given points changes with out of plane rotations.  A second drawback is that differences in depth alone cannot account for local point properties.  
% ; for example Such global properties cannot cope with transformations
%to cope with transformations.  For example, the depth difference 
%properties such as the direction of the local surface normal, into consideration. but all of them ignore the local surface normal direction. %\CW{this resembles the common part based model assumption, that local patch always more consistent against the global one}
%The determination of two offsets are either defined \textit{w.r.t.} the camera frame as in \cite{Shotton2013_pami} or under an assumed local frame \cite{xuHand_13,cascaded-hand} that don't utilize the local surface normal direction.
%\AY{expand on weakness here, even better if we can give image examples}\CW{some examples in feature_compare.eps, but not very obvious}

%Our random normal difference feature follows the same formalism as the random depth difference feature proposed in  that is used in most other random forest based pose regression frameworks. For simplicity, we use a more general term \textit{random difference feature} to indicate both types of features where we discuss them in a joint context.
%the proposed random normal difference feature in this paper and conventional random depth difference feature. 
%We first formulate the random difference feature.  We then provide two necessary conditions for 3D rigid transformation invariance and detail our normal difference feature. 

\subsection{Pose conditioned random normal difference feature}
\label{sec:ndf}
Surface normals are an important local feature for many point-cloud based applications such as registration~\cite{Rusu-pfh} and object detection~\cite{Rusu-kitchen,linemod,dl-rgbd}. Surface normals would seem a good cue for hand pose estimation too, since the direction of the surface helps to establish the local reference frame, as will be described in~\ref{sec:rf}.  For two given points, the angular difference between their normal directions remains unchanged after rigid transformations. Hence, we propose a pose-conditioned  
normal difference feature which is highly robust towards 3D rigid transformations.

To make random features invariant to 3D rigid transformations \textit{i.e.}, 
\begin{equation}
f_{\mathcal{I}}(\mathbf{p}_i, \mathbf{\delta}_1, \mathbf{\delta}_2) = f_{\mathcal{I}'}(\mathbf{p}'_i, \mathbf{\delta}_1, \mathbf{\delta}_2),
\label{equ:invariant}
\end{equation}
\noindent where $\mathcal{I'}$ and $\mathbf{p'}_i \in \mathcal{R}^3$ are the depth map and point position after transformation, it is necessary to satisfy the following two conditions: 

\begin{description}
   \item[i] The random offset generator $r(\cdot, \cdot)$ should be invariant to rigid transformations, \ie 
   \begin{equation}
       T(r(\mathbf{p}_i, \mathbf{\delta}_j)) = r(T(\mathbf{p}_i), \mathbf{\delta}_j),
   \end{equation}
where $T(\mathbf{q}) = \mathbf{R}\cdot \mathbf{q} + \mathbf{t}$ is the rigid transformation with $\mathbf{R} \in \mbox{SO(3)}$\footnote{Readers unfamiliar with Lie group matrix notations may refer to http://ethaneade.com/lie.pdf for more details. In short, SO(3) represents a 3D rotation while SE(3) represents a 3D rigid transformation.} and $\mathbf{t}$ as its rotation and translation respectively. This condition is equivalent to guaranteeing that the relative position between $\mathbf{p}_i$ and $r(\mathbf{p}_i, \mathbf{\delta}_j)$ remains unchanged after transformation, \textit{i.e.}, $ T(\mathbf{p}_i - r(\mathbf{p}_i, \mathbf{\delta}_j)) = T(\mathbf{p}_i) - r(T(\mathbf{p}_i), \mathbf{\delta}_j)$.
   %The relative  \CW{Left of the equation means the random offset firstly generated and then transformed, while right of the equation means the random offset directly generated after transformation.} This condition is equivalent to $T(r(p_i, \delta_j)) = r(T(p_i), \delta_j)$, which means $r()$ should be invariant to rigid transformation.
   %and $r(T(p_i), \delta_j)$ represents the random position calculated after transformation;~\AY{associative property?}\CW{like the linear or affine invariance for a function}
   
   \item[ii] The feature difference $\Delta(\cdot, \cdot)$ should be invariant to rigid transformation, \ie
   \begin{equation}
        \Delta(\phi_{\mathcal{I}}(\mathbf{q}_1), \phi_{\mathcal{I}}(\mathbf{q}_2)) = \Delta(\phi_{\mathcal{I'}}(\mathbf{q'}_1), \phi_{\mathcal{I}}(\mathbf{q'}_2)), 
   \end{equation}
\noindent 
where $\mathbf{q}_j'= T(\mathbf{q}_j), j\in\{1,2\}$ is the transformed offset position.
   %~\AY{what is the relationship of $q'$ to $q$ here?  explicitely state transformation T as previous }
\end{description}

\noindent 
%It's obvious that all previous random feature\cite{Shotton2013_pami,xuHand_13,cascaded-hand} is invariant to translation. We now narrow down our discussion to only rotation transformation.
To meet condition \textbf{i}, we extend the random position calculation $r(\mathbf{p}_i, \mathbf{\delta}_j)$ as 
\begin{equation}
\label{equ:offset}
r(\mathbf{p}_i, \mathbf{\delta}_j, \mathbf{R}_i) = \mathbf{p}_i + \mathbf{R}_i\cdot \mathbf{\delta}_j, 
\end{equation}
\noindent
where $\mathbf{R}_i \in \mbox{SO(3)}$ is a latent variable
%~\AY{why doesn't $C_i$ does not appear in this equation?} 
representing the pose of local reference frame~\ref{sec:rf}.  For any rigid transformation $\mathbf{T} =  
        \begin{bmatrix} 
        \overline{\mathbf{R}} & \overline{\mathbf{p}} \\ 
        0 & 1 \end{bmatrix}$, 
Equ. \ref{equ:offset} satisfies condition \textbf{i} \textit{iff}
\begin{equation}
\label{equ:cond1}
    \mathbf{R}'_i =  \overline{\mathbf{R}}\mathbf{R}_i,
\end{equation}
where $\mathbf{R}_i$ and $\mathbf{R}'_i$ are the estimated latent variable before and after rigid transformation respectively.  In comparison to~\cite{cascaded-hand}, which also uses a latent variable $\mathbf{R}$, the $\mathbf{R}$ is estimated globally and therefore can be sensitive to the initialization. For us, the local Darboux frame is established through the local surface normal direction (see Section~\ref{sec:regression}) and has no such sensitivity.  

To meet condition \textbf{ii}, given the random positions $\mathbf{q}_1$ and $\mathbf{q}_2$, we use the direction of the normal vector as our local feature map. The feature difference is cast as the angle between two normals, \ie
\begin{equation}
    \Delta(\phi_{\mathcal{I}}(\widetilde{\mathbf{q}_1}), \phi_{\mathcal{I}}(\widetilde{\mathbf{q}_2})) = n(\widetilde{\mathbf{q}_1})\cdot n(\widetilde{\mathbf{q}_2}),
\end{equation}
\noindent
where $\widetilde{q}$ denotes the 2D projection of the random position onto the image plane, since the input 2.5D point cloud is indexed by the 2D projection coordinates. $n(\cdot)\in \mathcal{R}^3$ denotes the corresponding normal vector. Since the angle between two normal vectors remains unchanged under a rigid transformation for any two given surface points, our feature also fulfills condition \textbf{ii}.  In comparison, the depth difference feature, as used in~\cite{cascaded-hand,xuHand_13,Shotton2013_pami}, does not fulfill this condition.

Our proposed normal difference feature can be computed based on any surface normal estimate.  We describe a conventional method based on eigenvalue decomposition in~\ref{sec:normal:accu} and then propose an efficient approximation alternative in~\ref{sec:normal:appro}.   %based on random forests in~\ref{sec:normal:appro}.  

%In contrast to the depth difference, our normal difference is much better preserved under 3D rigid transformations and articulation changes.  

%Our normal difference feature better handle the 3D rigid transformation than previous method since 1) while \cite{cascaded-hand} is 
%~\AY{the cascaded paper is also using the latent variable C, which affords them some invariance} which is also conditioned on the previously estimated frame.    
%which also takes the random difference feature conditioned on an estimated pose $R'$.  We have two major differences to \cite{cascaded-hand}, 1) while in \cite{cascaded-hand} $R'$ is estimated in a cascaded manner by starting from an initial guess $R'_0$ and progressively updating the current guess to get $R'_t$ after $t$ iterations, we utilize the normal direction of local surface to directly estimate $R'$; this helps to avoids the large errors from poor intializations; 2) we use a normal difference feature which meets condition \textbf{ii} instead of the depth feature used in \cite{cascaded-hand}.

\begin{figure}[b]
    \centering
    \includegraphics[width=1.\textwidth]{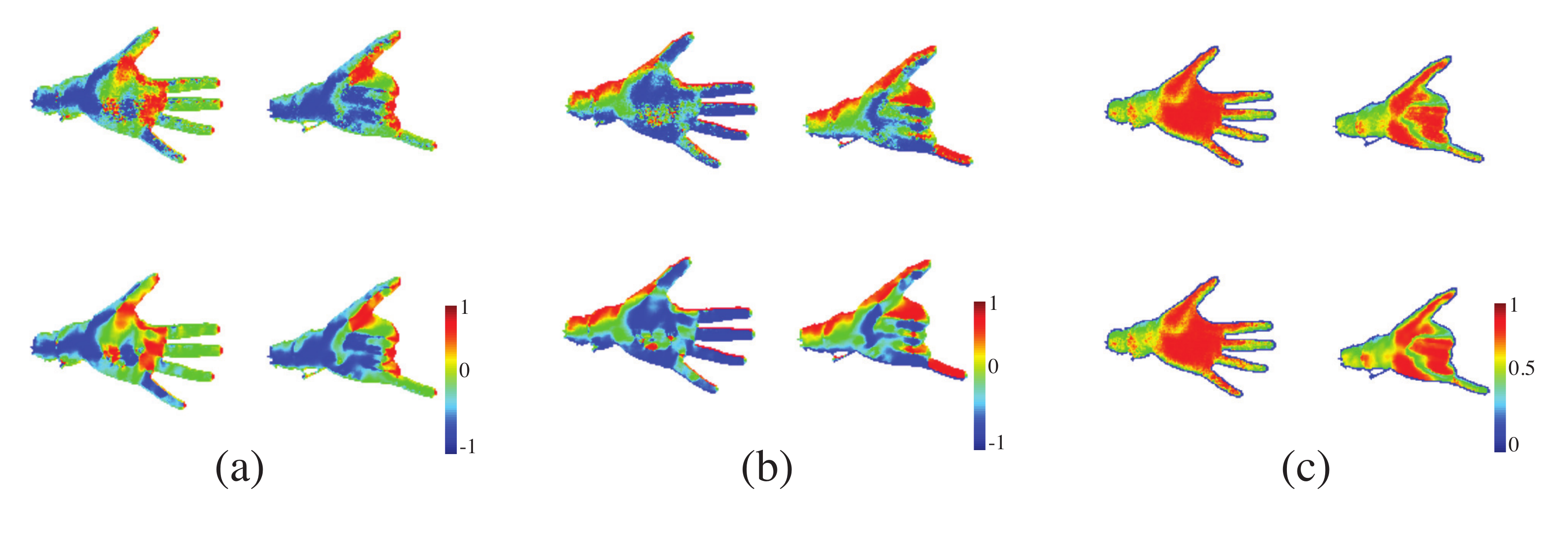}
    \caption{Estimated surface normal. From (a) to (c) the x, y, z-axis coordinate of the normal vector, resp. The first row is the regressed surface normal by the random forest and the second row is estimated by PCA. \textbf{(Best viewed in colour)}}
    \label{fig:normal}
\end{figure}

%\CW{according to my later experiments and visualization, discrimination, i.e., to distinguish minor finger pose difference, especially in the twisted case, is doubtful, compared to depth difference feature. Thus maybe in this work we shall just limit our motivation to its invariance properties to rigid body transformation; its discrimination by firstly estimate the feature map by normal estimation, then performing regression upon that and explicitly handle local/contextual properties are ignored.} 

%In Section \cite{ndf} we introduce the proposed frame conditioned random normal difference feature.  To this end, we justify that the proposed feature well handles 3D rigid transformation and at the same time keeping the efficiency of random forest. 

\subsection{Surface normal estimation based on eigenvalue decomposition}
\label{sec:normal:accu}

For an input 2.5D point cloud, we distinguish between inner points that lie inside the point cloud and edge points on the silhouette of the point cloud.  For edge points, normal estimation degenerates to 2D curve normal estimation since the normal direction is constrained to lie in the image plane. 

For inner points, the local surface can be approximated by the $k$-neighbourhood surface direction\cite{Rusu-kitchen}.  The eigenvector corresponding to the smallest eigenvalue of the neighbourhood covariance matrix can be considered the normal direction.  The sign of the normal direction is further constrained to be the same as the projection ray.  In our implementation, we set $k$ as 10 mm. We show estimated normals in the second row of Fig. \ref{fig:normal}.  Our preliminary experiments show that $k$ values from 5mm to 15mm all return comparable normal estimates; above 15mm, performance starts to deteriorate, presumably due to points from multiple fingers being grouped together into the same neighbourhood.

% and readers can look up to \cite{Rusu-kitchen} for more details.

% By taking PCA on the neighbour covariance matrix,  

\subsection{Surface normal regression with random forests}
\label{sec:normal:appro}
% The accurate estimation of normals requires eigen-decomposition at every single inner point. this is computationally expensive.

Estimating the normal at every inner point in the point cloud can become very computationally expensive, with an eigenvalue decomposition per point.  Alternatively, we can take advantage of the efficiency of random forests and regress an approximate normal direction.  
%Random forests are ensembles of decision trees. The trees are composed of split nodes associated with weak learners to make decisions based on only one feature dimension, and leaf node store the learned regression values. Readers are referred to \cite{Shotton2013_pami} for more details.
Directly regressing the normal vectors in vector space does not maintain unit length %property of normal vector, 
so we parameterize the normal vector with spherical coordinates $(\theta, \varphi)$ where $\theta$ and $\varphi$ are the polar and azimuth angles, resp. $\theta$ and $\varphi$ are independent and can be regressed separately. %Similar to~\cite{biternion}, 
We model the distribution of a set of angular values $\mathcal{S} = \{\theta_1, \cdots \theta_n \}$ as a Von Mises Distribution, which is the circular analogue of the normal distribution.  The distribution is expressed as 
\begin{equation}
        p_{VM}(\theta_i | \mu, \kappa) = \frac{e^{\kappa cos(\theta_i-\mu)}}{2\pi I_0({\kappa})},
        \label{equ:vm}
\end{equation}
\noindent
where $\mu$ is the mean of the angles, $\kappa$ is inversely related to the variance of the approximated Gaussian and $I_0({\kappa})$ is the modified Bessel function of order 0.
%\cite{biternion}. 
To estimate the mean and variance of the distribution, we first define
\begin{equation}
   \overline{C} = \sum_{i}\cos(\theta_i),~ 
   \overline{S} = \sum_{i}\sin(\theta_i),~
   \overline{R} = (\overline{C}^2 + \overline{S}^2)^{\frac{1}{2}}.
\end{equation}
\noindent
Then the maximum likelihood estimates of $\mu$ and $\kappa$ are
\begin{equation}
    %\begin{split}
    %&\mu = \mbox{atan2}(\overline{S}, \overline{C}) \\       
    %&\overline{R} = \frac{I_1({\kappa})}{I_0({\kappa})}.
    \mu = \mbox{atan2}(\overline{S}, \overline{C}) \qquad \text{and} \qquad
    \overline{R} = \frac{I_1({\kappa})}{I_0({\kappa})}.
    %\end{split}
\end{equation}
During training, each split node is set by maximizing the information gain as
\begin{equation}
    I = H(\mathcal{S}) - \sum_{i\in\{L,R\}}\frac{|\mathcal{S}^i|}{|\mathcal{S}|}H(\mathcal{S}^{i}),
\end{equation}
\noindent
where the entropy of the Von Mises Distribution is defined as 
\begin{equation}
    H(\mathcal{S}) = \mbox{ln}(2\pi I_0(\kappa)) - \kappa\frac{I_1({\kappa})}{I_0({\kappa})}.
\end{equation}

The training procedure for the random forest that estimates the normal is almost identical to~\cite{Shotton2013_pami} with the exception that the random offsets are restricted to lie within the region of the same $k$-nearest neighbourhood that was used for the eigenvalue decomposition based normal estimation in~\ref{sec:normal:accu}. The mean of the angular values propagated to each leaf node is selected as the leaf node's prediction value. In practice, to make the normal regression even more efficient, we combine the estimation of $\theta$ and $\varphi$ into one forest by regressing the $\theta$ in the first 10 layers and $\varphi$ in the later 10 layers, rather than estimating them independently.  

Since the random offset is limited to a small area, which restricts the randomness of the trees, we find that the average error between approximated and true normal directions only goes up from~\mytilde12\degree{} to~\mytilde14\degree{} when decreasing the number of trees from 10 to 1. As the normal difference feature is not sensitive to such minor errors, we use only 1 tree for all experiments in this paper. The proposed method is extremely efficient; normals for input point clouds can be estimated in \mytilde4 ms on average, compared to \mytilde14 ms based on eigenvalue decompositions on the same machine.

\section{Frame conditioned regression forest}
\label{sec:rf}
We formulate hand joint estimation as a regression problem by regressing the 3D offsets between an input 3D point and a subset of hand joints. Directly regressing all joints of the hand at once, as has been done in previous works~\cite{forth-oneshot,Keskin2012-ns} is difficult, given the highly articulated nature of the hand and the many ambiguities due to occlusions and local self-similarities of the fingers.  Instead, we prefer to solve for the joints in a hierarchical manner, as  state-of-the-art results~\cite{cascaded-hand,Tang2015-blackbox} have demonstrated the benefits of solving the pose progressively down the kinematic chain.  

%is insufficient to handle the large pose variance and always causes large error. 

In this section, we propose a conditional regression forest, namely the \textit{Frame Conditioned Regression Forest} (FCRF) which performs  regression conditioned on  information estimated in the previous stages. The hand joints are regressed hierarchically by following the kinematic chain from wrist down to the finger joints.  
%according to the tree-structured topology of hand and the 
At each stage, we first estimate the reference frame based on results of previous stages and then regress the hand joints relevant to that stage with the FCRF.  % is estimated based on results from earlier stages
%Estimation at each stage is conducted by firstly estimating the frame based on the results from earlier stages and regressing the hand joints for current stage via FCRF based on the estimated frame,  The formulation of FCRF together with its training and testing algorithms are provided in this section. The hierarchical regression scheme and frame estimation method for each stage will be detailed in Section \ref{sec:regression}.

There are three main benefits to using the FCRF.  First of all, offsets between input points and finger joints are transformed into the local reference frame. This reduces the variance of the offsets and simplifies the training.  It also implicitly incorporates skeleton constraints provided by the training data.  Secondly, the related normal difference feature, as described in section \ref{sec:normal}, is conditioned on the estimated reference frame and makes the joint regression highly robust to 3D rigid transformations.  Finally, 
%trees from different stages can be trained simultaneously. 
FCRF is in-plane rotation-invariant, and does not need manually generated in-plane rotated training samples for training as in \cite{cascaded-hand,Tang-LRF,Li2015-sip}, so the training time and resulting tree size can be reduced significantly.
%\AY{last point weak --> substantiate why}\CW{\cite{Li2015-sip}\cite{Tang-LRF}'s work cannot, while \cite{cascaded-hand}'s work can} 

Specifically, given input point $\mathbf{p}_i \in \mathcal{R}^3$ from the point cloud, the FCRF for the $j^\text{th}$ stage solves the following regression
\begin{equation}
\label{equ:fcrf}
    \mathbf{O}_{j}^{(i)} = r_{j}(\mathcal{I}, \mathbf{C}_{j}^{(i)}),
\end{equation}
\noindent 
where $\mathbf{O}_{j}^{(i)} \in \mathcal{R}^{3\times n}$ is the offsets between input point $\mathbf{p}_i$ and the $n$ joints to be estimated in $j^\text{th}$ stage, $\mathcal{I}$ denotes the input depth map and $\mathbf{C}_{j}^{(i)} \in \mbox{SE(3)}$ is the corresponding local frame. We define the position of the input point $\mathbf{p}_i$ as the origin of the local reference frame, \ie
\begin{equation}
\label{equ:frame}
    \mathbf{C}_{j}^{(i)} = 
    \left[
        \begin{array}{c|c} \mathbf{R}_{j}^{(i)} & \mathbf{p}_i \\ 
         \hline 0 & 1 \end{array}
    \right],
\end{equation}
\noindent
where $\mathbf{R}_{j}^{(i)} = \begin{bmatrix}\mathbf{x}, \mathbf{y}, \mathbf{z}\end{bmatrix}\in \mbox{SO(3)}$ is a rotation matrix representing the frame pose, and $\mathbf{x},\mathbf{y},\mathbf{z}\in \mathcal{R}^3$ are the corresponding axis directions. Both $\mathbf{R}_i$ and $\mathbf{p}_i$ are defined with respect to the camera frame. 
%\AY{R is a rotation matrix, cannot be equal to $[x,y,z]$}

The regression $r_j(\mathcal{I}, \mathbf{C}_{j}^{(i)})$ is done by a random forest. 
%Random forests are ensembles of decision trees. The trees are composed of split nodes associated with weak learners to make decisions based on only one feature dimension, and leaf node store the learned regression values. Readers are referred to \cite{Shotton2013_pami} for more details.
%{\bf LUC: IT IS STRANGE THAT AFTER HAVING USED THE TERM RANDOM FOREST SO OFTEN BEFORE, IT THEN ALL OF A SUDDEN GETS EXPLAINED... PROBABLY THE EXPLANATION CAN BE REMOVED... OR PLACED EARLIER.}

During training, $\mathbf{o}_{ik} \in \mathcal{R}^3$, the offset between point $\mathbf{p}_i$ and joint $\mathbf{l}_k$ to be estimated, is first rotated to the local reference frame $\mathbf{C}_j^{(i)}$ as $\widetilde{\mathbf{o}_{ik}}$, \ie
\begin{equation}
    \widetilde{\mathbf{o}_{ik}} = (\mathbf{R}_j^{(i)})^{T}\cdot \mathbf{o}_{ik}.
\end{equation}
The distribution of offset samples are modeled as a uni-modal Gaussian as in \cite{Shotton2013_pami}. For each split node of the tree, the normal difference feature which results in the maximum information gain from a random subset of features is selected. For each leaf node, mean-shift searching~\cite{meanshift} is performed and the maximal density point is used as the leaf prediction value. 
%\AY{details on random forests a bit too sparse, should be written as if reader does not know random forests.}

During testing, given the estimated local frame $\mathbf{C}_j^{(i)}$, the resulting offset $\mathbf{o}_{ik}$ can be re-projected to the camera frame as 
\begin{equation}
   \mathbf{o}_{ik} = (\mathbf{R}_j^{(i)})\cdot \widetilde{\mathbf{o}_{ik}}.
\end{equation}

\section{Hierarchical hand joint regression}
\label{sec:regression}
In this section, we detail the design of reference frames used by FCRFs in every stage, given the estimated local surface normal and the parent joint positions from previous stages. Free moving hand pose estimation faces two major challenges, \textit{i.e.}, large variations of viewpoints, and self-similarities of different fingers. We decompose hand pose estimation into two sub-problems that explicitly tackle these two challenges: first, we estimate the reference frame of the palm and second, we estimate the finger joints. %Specifically, we firstly distinguish the different local properties of edge point and inner point on the point cloud to use different point with different local properties for estimation in different stages(only use edge for the wrist center localization). Then we use the normal direction of local surfaces to establish the Darboux local frame for conditional regression.

In Sections \ref{sec:rf_wrist} and \ref{sec:rf_mcp} the palm estimation is introduced by first estimating the wrist joint (palm position) followed by MCP joints(Fig.~\ref{fig:framework}(a))
%{\bf LUC: IT WOULD SEEM TO ME THAT YOU HAVE NEVER BEFORE EXPLAINED WHAT MCP MEANS.} 
for all 5 fingers (palm pose), in which the Darboux frame for every input point is established by taking the estimated wrist joint as reference point. In Sections \ref{sec:rf_pip} and \ref{sec:rf_tip} the joints for each finger are estimated, progressively conditioned on the previously estimated joint position. 
%For the rest of the paper, we use the term wrist and palm interchangeably to denote the root part of the hand.  
%{\bf LUC: USING WRIST AND PALM INTERCHANGEABLY IS STRANGE THOUGH, AS THE PALM CAN DEFINITELY ROTATE WITH RESPECT TO THE WRIST.}
%We note that, %It should also be noted that different terminology
%due to a different parametrization of the hand skeleton, estimating the child joint position in our work is equivalent to estimating the parent joint angle in ~\cite{cascaded-hand,Tang2015-blackbox}.
% {\bf LUC: I DO NOT COMPLETELY UNDERSTAND THE ABOVE STATEMENT.}\CW{we are estimating the child joint position, this is same as estimaing the rotated joint angle of the parent joint(there are some discriminative work estimating the joint angle, instead of position)}
\subsection{Wrist estimation}
\label{sec:rf_wrist}
We consider only edge points on the hand silhouette as inputs for estimating the wrist joint. Our rationale is that we cannot find unique reference frames for non-edge points, since knowing only the direction of the normal, \ie the z-axis, is insufficient to uniquely determine the x- and y-axis on the tangent plane. We assume orthographic projection for the point cloud, \ie the tangent plane of edge point is orthogonal to the image plane, then the local reference frame of edge point $\mathbf{p}_i$ can be defined uniquely as follows,
%For edge points, the local reference frame can be defined uniquely, since it degenerates to the case of estimating the normal vector of a 2D curve. Specifically, the local reference frame of edge point $p_i$ is defined as
\begin{equation}
\label{equ:wrist}
\begin{split}
    &\mathbf{x}_{wrist}^{(i)} = \mathbf{n},\\
    &\mathbf{y}_{wrist}^{(i)} = \mathbf{z}_{wrist}^{(i)} \times  \mathbf{x}_{wrist}^{(i)},\\
    &\mathbf{z}_{wrist}^{(i)} = \mathbf{n}_i,
\end{split}
\end{equation}
\noindent
where {\bf n} is the image plane normal direction, ${\bf n}_i$ is the normal to the silhouette at point $i$. 
%{\bf LUC: THIS DERIVATION DIFFERS FROM WHAT I WOULD HAVE EXPECTED: Ni IS THE NORMAL TO THE SILHOUETTE, A SECOND DIRECTION WOULD BE THE TANGENT TO THE SILHOUETTE, THE NORMAL TO THE IMAGE PLANE FOLLOWS AS THE CROSS PRODUCT. FROM WHAT YOU NOW SAID, THIS WOULD IMPLY AN ASSUMPTION OF ORTHOGRAPHIC PROJECTION... THAT WOULD BE IMPORTANT TO STATE EXPLICITLY.}\CW{here I assume the projection ray is in the tangent plane, which is not exactly the case, but approximately. We uniformly use the axis order as x->y->z, instead of the calculation order.}
The resulting local reference frame is not only invariant to 2D rotations in the image plane but to some degree also robust to out-of-plane rotations, provided that the hand silhouette does not change too much.

\subsection{Metacarpophalangeal (MCP) Joint Estimation}
\label{sec:rf_mcp}
Given the estimated wrist point position as a reference point, we assume its relevant position under the local frame $C_{MCP}^{(i)}$ is unchanged then the local reference frame for point $\mathbf{p}_i$ is established as follows
\begin{equation}
\label{equ:mcp}
\begin{split}
    &\mathbf{x}_{MCP}^{(i)} = \mathbf{y}_{MCP}^{(i)}\times \mathbf{z}_{MCP}^{(i)}, \\
    &\mathbf{y}_{MCP}^{(i)} = \frac{\mathbf{n}_i\times(\mathbf{p}_{wrist}-\mathbf{p}_i)}{\|\mathbf{n}_i\times(\mathbf{p}_{wrist}-\mathbf{p}_i)\|_{2}}, \\
    &\mathbf{z}_{MCP}^{(i)} = \mathbf{n}_i,
\end{split}
\end{equation}
\noindent
where the z-axis of the local reference frame is defined as the normal direction $\mathbf{n}_i$, and the y-axis is defined by taking the wrist location $\mathbf{p}_{wrist}$ as a reference point. 
%{\bf LUC: IT IS NOT CLEAR TO ME HOW A POINT (WRIST LOCATION) FIXES A DIRECTION. OR IS IT RATHER THE LINE CONNECTING Pi AND Pwrist THAT IS USED? THAT IS WHAT THE FORMULA SEEMS TO SAY.}\CW{given the wrist location as a reference point, we assume that for a certain point p_i, its relative position to wrist point is unchanged, i.e. the normalized coordinate of wrist point in p_i's local frame is unchanged.}
The MCP joints from all five fingers are then regressed simultaneously, \textit{i.e.}, $\mathbf{O}_{MCP}^{(i)} \in \mathcal{R}^{3\times 5}$ using our previously defined FCRF.  

The estimated MCP joints are then replaced by the transformed MCP position from a template palm to reduce the accumulated regression error.
%~\AY{sounds hacky -- why do you need to replace?}\CW{not really, we replace by the template to reduce the regression error, like 5 mcp, if 4 of them are regressed correctly, but one not, then the palm frame will still not be influenced too much, but by replacing with the template, we can reduce the error of the wrong one}
We first find a closed form solution of the palm pose using a variation of ICP~\cite{aicp}.  The palm pose matrix $\mathbf{R}_{palm}$'s y-axis is defined as the direction from the wrist to the MCP joint of the middle finger, the z-axis is defined as the palm normal.
% \AY{define in terms of wrist / MCP equations; is $R$ a rotation matrix or an $xyz$ coordiante?  Not really interchangeable...}

\subsection{Proximal Interphalangeal (PIP) Joint Estimation}
\label{sec:rf_pip}
In the estimation of the PIP joint for finger $k$, all input reference frames share the same pose as the rotated palm reference frame as follows,
\begin{equation}
    \mathbf{C}_{PIP_{k}}^{(i)} = 
    \left[
        \begin{array}{c|c} \mbox{Rot}_k(\mathbf{R}_{palm}) & \mathbf{p_i} \\ 
         \hline 0 & 1 \end{array}
    \right],
    \label{equ:pip}
\end{equation}
\noindent
where $\mbox{Rot}_k(\cdot)$ is an in-plane rotation to align the reference frame's y-axis to the $k-^\text{th}$ finger's empirical direction Fig.~\ref{fig:framework} (a).
%\AY{is the direction PIP - MCP?  Doesn't this imply already knowing the PIP?  or can this be defined wrt MCP?}\CW{it's a bit like the mean of the finger direction distribution.}.

Given the local self-similarity between fingers, it can be easy to double-count evidence. To avoid this, we adopt two simple measures.  First, we use points only from the neighbourhood of the parent MCP joint as input for regressing each PIP joint, since these points best describe the local surface distortion raised by the parent joint articulation~\cite{Kovalsky-learnArticulation}.
%~\AY{what does the distortion mean?}\CW{different joint angle of parent joint, rises different types of distortion on that local area}.  Another way considering the problem is that estimating the location of the child joint position is equivalent to estimating the transformation of the parent joint. 
Secondly we limit the offset of the FCRF to lie along the direction of the finger to maintain robustness to noisy observations from nearby fingers.  
%makes the feature selection in random forest training more generalized.~\AY{how is this more generalized?}~\CW{avoid the RF overfitted to some noisy feature difference that might be evident from other fingers}

\subsection{Distal Interphalangeal Joint (DIP) and Finger Tip (TIP) Estimation}
\label{sec:rf_tip}
The ways to estimate DIP and TIP joints are identical, since their parents are both 1-DoF joints. %The poses of local frame for estimating one certain joint $l$ is the same for different input point. 
The local reference frame for each joint is defined as follows
\begin{equation}
\begin{split}
    &\mathbf{x}_l= \mathbf{z}_{palm}\times \mathbf{y}_l,\\
    &\mathbf{y}_l = \mathbf{p}(l)-\mathbf{g}(l), \\
    &\mathbf{z}_l = \mathbf{x}_l\times \mathbf{y}_l,
\end{split}
\end{equation}
\noindent
where $\mathbf{z}_{palm}$ is the normal direction of palm, $\mathbf{p}(l)$ and $\mathbf{g}(l) \in \mathcal{R}^3$ denote the parent and grandparent joint of $l$ respectively. To avoid double counting of local evidence, we adopt the same techniques as in section~\ref{sec:rf_pip} .

%\noindent\textbf{Discussion}
% \subsection{Alternative Regression Schema}
% Note that we are not directly using the skeleton model of the hand for regression; however, as the regressed offset is aligned  \textit{w.r.t.} the input reference frame, we are implicitly encoding skeleton constraints into the regression framework.  \AY{this point separate from following points, consider placing elsewhere; move alternative schema to experimental results}

% We have found regressing the offset of input points $p_i$ to each finger joint was the most robust scheme.  In our preliminary experiments, we also tried directly regressing each finger joint's joint angle, as well as the offset between parent and child joints, using the same technique as described above.  Both strategies showed inferior results (see Section \ref{sec:exp}).  We speculate that these regression schemes are less robust since the position of child joint is sensitive to error from the estimated parent joint's reference frame~\AY{isn't this still true?!}.  

% For a single training depth map, different input points are corresponding to one single predicted offset, which limits the randomness of the random forest.~\AY{explanation here weak?  not sure what you mean by this still}  Notice that this is also the case for the first strategy.

\section{Experiments}
\label{sec:exp}
\begin{figure}[!htbp]
    \centering
    \includegraphics[width=.95\linewidth]{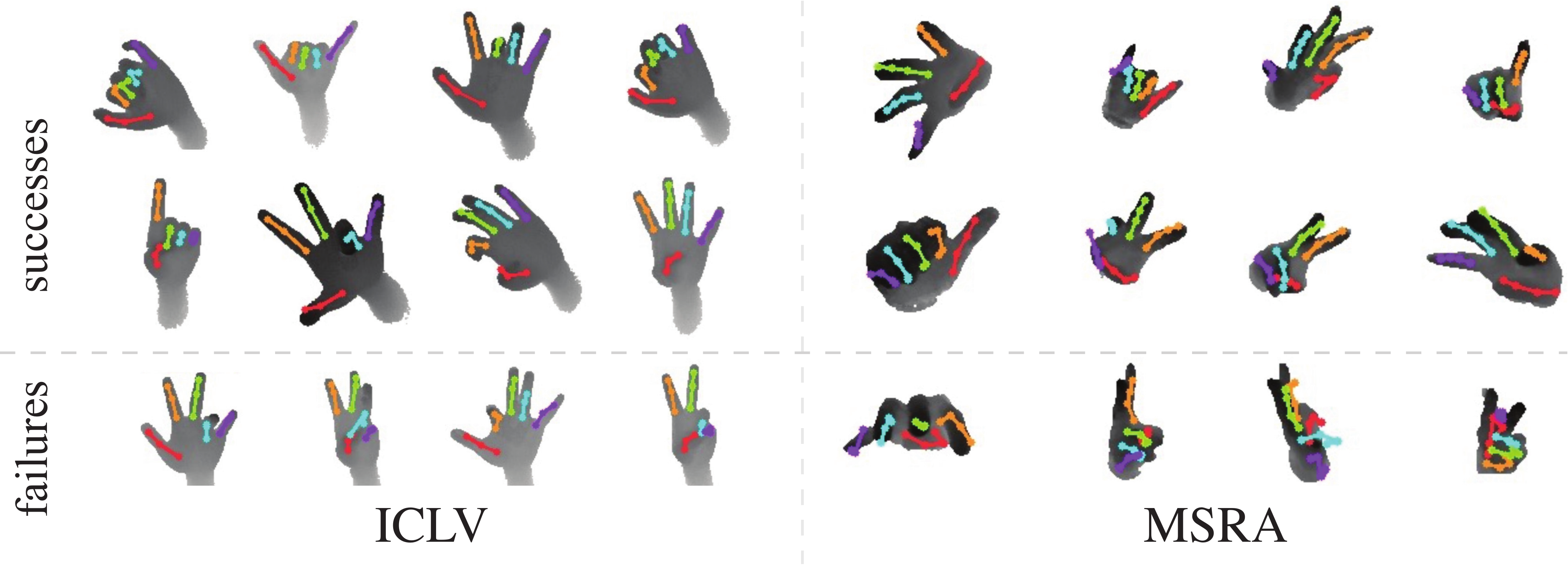}
    \caption{Examples of successful and failed pose estimates on the ICLV~\cite{Tang-LRF} and the MSRA~\cite{cascaded-hand} dataset.  Failures are due to extreme view point, wrongly estimated normal direction, etc. (best viewed in colour)}
    \label{fig:qual}
    \vspace{-1.0cm}
\end{figure}

We apply our proposed hand estimation method to two publicly available real-world hand pose estimation datasets: ICLV~\cite{Tang-LRF} and MSRA~\cite{cascaded-hand}.  The performance of our method is evaluated both quantitatively and qualitatively. For quantitative evaluation, two {evaluation metrics}, per-joint error (in mm) averaged over all frames and percentage of frames in which all joints are below a threshold~\cite{Vitruvian}, are used.  We show qualitative results in Fig.~\ref{fig:qual} and encourage the reader to watch the accompanying supplementary videos. %~\AY{website, disclosed after publication}.

All experiments are conducted on an Intel 3.40 GHz I7 machine and the \textit{average run time} is 29.4fps or 33.9ms per image. The \textit{maximum depth} of all the trees is set to 20. The \textit{number of trees} for all joint regression forests are set to 5 and 1 for normal estimation (see Section \ref{sec:normal:appro}). 

To highlight the effectiveness of our proposed normal difference feature, we first apply our frame conditioned regression forests with the same hierarchical structure but based on the standard depth difference feature~\cite{Shotton2013_pami}.  We denote this variation using the depth difference feature as our \textit{baseline method}.  It should be noted that the baseline does depend on normal estimation for the establishment of the local wrist frame. We also compare to methods directly regressing the wrist and MCP joint positions without establishing the frame~\cite{Tang-LRF,Li2015-sip} or based on an initial guess and the subsequent, iterative regression of the error~\cite{cascaded-hand}.

\subsection{ICLV hand dataset}
The ICLV hand dataset~\cite{Tang-LRF} has 20K images from 10 subjects and an additional 160K in-plane rotated images for training.  Since our method is invariant to in-plane rotation, we train with only the initial 20K.  %In comparison to~\cite{Tang-LRF,cascaded-hand,Li2015-sip}, which also use random forests, this allows us to not only reduce training time, but  need to be trained on the whole dataset, ours significantly reduces the total training samples size by 8 times which further lead to reduced training time and resulted tree size.
The test set is composed of 2 sequences with continuous finger movement but little viewpoint change. 
%} 

\begin{figure}[!htbp]
    \centering
    \includegraphics[width=1.0\textwidth]{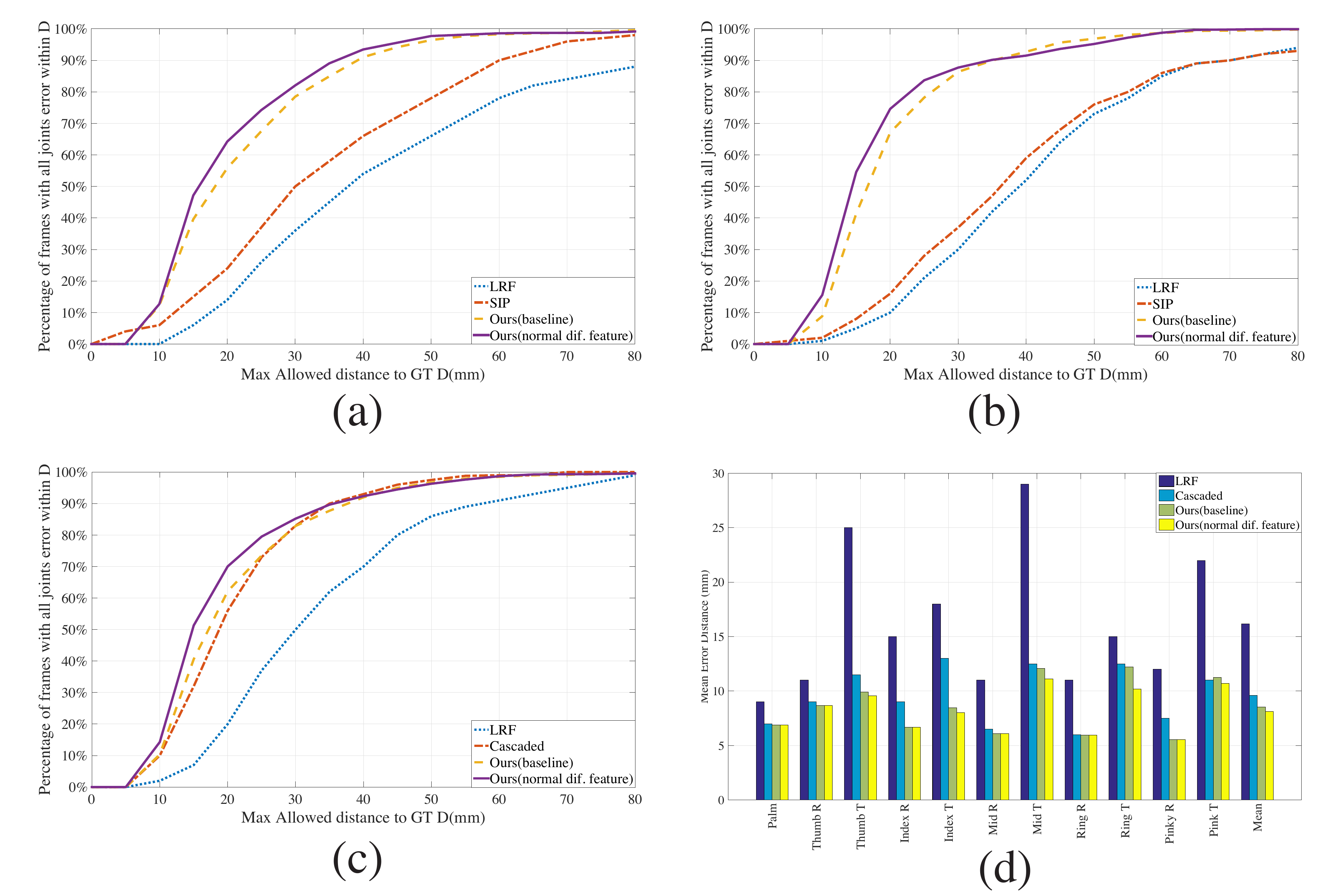}
    \caption{Quantitative evaluation on ICLV dataset. From (a) to (c), success rates over different thresholds on sequence A, B and both respectively. (d) pre-joint average error on both sequences (R:root, T:tip)}
    \label{fig:icl_res}
    \vspace{-0.5cm}
\end{figure}

We compare our method (both the baseline and the version with the normal difference feature) against the state-of-art methods Latent Regression Forest (LRF)~\cite{Tang-LRF}, Segmentation Index Points(SIP)~\cite{Li2015-sip}, and Cascaded Regression (Cascaded)~\cite{cascaded-hand}.  Fig.~\ref{fig:icl_res}(a)-(c) shows that both variations of our proposed method outperform LRF~\cite{Tang-LRF} and SIP~\cite{Li2015-sip} by a large margin on both test sequences.  In comparison to the Cascaded method of~\cite{cascaded-hand}, shown in Fig. \ref{fig:icl_res}(c), our baseline is comparable or better at almost all allowed distances, while the variation with the normal difference feature boosts performance by another $5-10\%$.  As shown in Fig.~\ref{fig:icl_res}(d),  our method significantly out-performs~\cite{Tang-LRF}, and it outperforms \cite{cascaded-hand} by \mytilde2mm in terms of the mean error.
%Among all finger joints, the largest mean error differences occur in index finger tip, which exhibits continuous finger movement in the test sequence, and shows our method can generalize to unseen poses compared to~\cite{cascaded-hand} which regress the finger as a whole.
These results confirm that conditioning finger localization on the wrist pose, as we have done and as is done in~\cite{cascaded-hand}, can significantly boost accuracy.  Furthermore, our proposed normal difference feature is able to better handle 3D rigid transformations.
% and distinguish finger positions with minor changes  (\AY{reference specific aspect of fig. d to prove this?}\CW{Maybe we just mention 3D rigid transformation is ok; otherwise by explaining the discrimination power of normal difference, makes it contradictory to the bad performance on MSRA dataset.}).

\subsection{MSRA hand dataset}
The MSRA hand dataset~\cite{cascaded-hand} contains 76.5K images from 9 subjects with 17 hand gestures. We use a leave-one-subject-out training/testing split and average the results over the 9 subjects. This dataset is complementary to the ICLV dataset since it has much larger viewpoint changes but limited finger movements. The sparse gesture set does not come close to reflecting the range of hand gestures in real-world HCI applications and as such, is not suitable for evaluating how well a method can generalize towards unseen hand gestures.  Yet, this dataset is very good for evaluating the robustness of pose estimation methods to 3D rigid transformations; for HCI applications, this offers flexibility for mounting the camera in different locations. %(strong correlation among different finger joints, which is not the case in real-world human-computer interaction gestures)

\begin{figure}[!htbp]
    \centering
    \includegraphics[width=1.\textwidth]{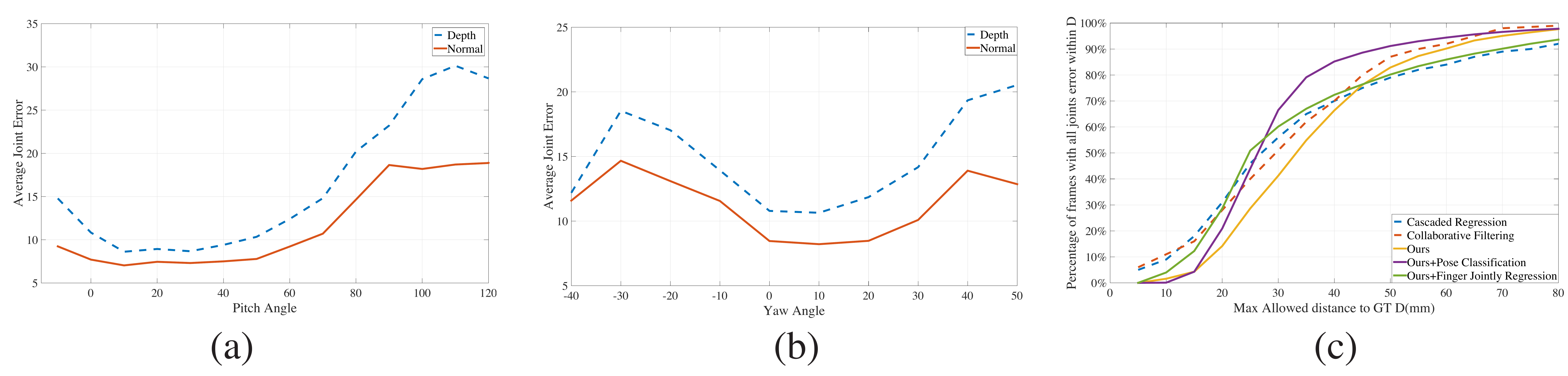}
    \caption{Quantitative evaluation on MSRA dataset. (a) to (b): average joint error as a function of pitch and yaw angle of the palm pose with respect to camera frame; (c) success rates over different thresholds.}
    \label{fig:msra_res}\vspace{-0.5cm}
\end{figure}
%{\bf LUC: THERE IS AN ERROR IN FIGURE (C): IT SHOULD BE `COLLABORATIVE FILTERING'}

As is shown in Fig.~\ref{fig:msra_res}(a)-(b), using the normal difference exhibits less variance to viewpoint changes than using the depth difference. This is more prominent in the pitch angle due to the elongated hand shape.  For a given pair of points, their depth difference exhibits larger variation \wrt pitch angle viewpoint changes. Nevertheless, the performance of the normal difference does decrease under large viewpoint changes. We attribute this to the errors in surface normal estimation due to point cloud noise and to the fact that a 2.5D point cloud only partially represents the full 3D surface.

We compare our proposed method against the state-of-the-art Cascaded Regression (Cascaded)~\cite{cascaded-hand} and the Collaborative Filtering (Filtering)~\cite{collatorative} approaches.  Above an allowed distance of 40mm to the ground truth, our approach is comparable to the others.  Below the 40mm threshold, our baseline and the normal difference feature version has around~\mytilde$14\%$ less frames than competing methods.  We attribute the difference to the fact that both the Cascaded and the Filtering approach consider the finger as a whole, in the former case for regression, and in the latter as a nearest neighbour search from the training data.  
%\CW{in msra dataset, configurtaion of the finger is sparse, onece MCP joint pose is determined, the rest two finger joints pose are also determined; for our initial result, we consider the 3 joints of a finger independently, so even our mcp joint pose estimation is correct, there's still great chance the rest two joints pose will be estimated wrongly, resulted into large error; while for cascaded hand case, by taking the finger as a whole, once the mcp joint pose is determined correctly, all the rest finger joints can be determined correctly.} 
While our method generalizes well to unseen finger poses by regressing each finger joint progressively, it is unable to utilize the sparse (albeit similar to testing) set of finger poses in the training. Nevertheless, in an HCI scenario, a user is often asked to first make calibration poses which are important to improve accuracy.  As such, we propose two minor modifications to make more comparable evaluations.

For the first modification, we first regress the palm pose, normalize the hand, and then classify the hand pose as a whole.  Based on the classification, we assign a corresponding pose sampled from the training set, transformed accordingly to the palm pose.  This modification, which we denoted as \textit{pose classification} is similar to Filtering~\cite{collatorative} as both methods consider the hand as a whole.  By classifying the 17 gesture classes as provided by the MSRA dataset we now outperform \cite{collatorative} over a large interval of thresholds larger than 22mm.  We attribute the increased performance to our accurate estimate of the palm pose.%invariance to 3D rigid transformations.

For the second modification, we regress each finger (\ie the 3 finger joints PIP, DIP, TIP) as a whole given the estimated palm pose. This is similar in spirit to the regression strategy in \cite{cascaded-hand} which takes each finger as a whole. Our method outperforms~\cite{cascaded-hand} by \mytilde$5\%$ in the 25-30mm threshold interval. We attribute this improvement to our palm pose estimation scheme which avoids sensitivity to initialization~\cite{cascaded-hand}. 

Despite our modifications, it should be noted that regressing the finger as a whole cannot generalize to unseen joint angle combinations for one finger, which is usually the case in real-world HCI scenarios, \eg grasping a virtual object, where one finger may exhibit various joint angle combinations according to the shapes of different objects. However, the two strategies are complementary, \ie regressing finger joints progressively can generalize to unseen finger poses while regressing the finger as a whole can capture finger joint correlations in training samples.  Given enough computational resources, the two strategies can be performed in parallel, with the best estimation being selected according to an energy function as in model-based tracking. We leave this as our future work.
%\AY{Must conclude this part on a positive note about the necessity of consdiering each joint individually... something about being unseen in the training data?}
%\CW{Given generalizing towards unseen finger poses is crucial, progressively regressing each joint independently . But regressing joint as a whole can otherwise capturing the dependencies among different finger joints. Given enough computational resources, the two finger regression strategies, \ie regressing finger joint indepedently and as a whole can be done parallelled, and choosing the best candidate }
%{\bf LUC: YOU COULD ADD SOMETHING LIKE: As generalizing towards unseen finger poses is crucial, we did not include the above modifications into our standard pipeline.  BUT IT REMAINS A PROBLEM THAT FOR 1 OUT OF 2 DATABASES THE STANDARD METHOD IS OUTPERFORMED BY THE COMPETITION... COULD WE COMBINE BOTH APPROACHES: FINGER DIGITS DEALT WITH SEPARATELY AND FINGERS DEALT WITH AS A WHOLE, THE SYSTEM DOES BOTH AND THEN AUTOMATICALLY CHOOSES THE BEST? THAT COULD MEAN HAVING TO RUN A MORE COMPLICATED SYSTEM, YET RANDOM FORESTS HAVE THE RIGHT STRUCTURE TO APPEND PARALLEL TREES FOLLOWING THE ALTERNATIVE APPROACH. THE ISSUE IS THEN FINDING A WAY TO AUTOMATICALLY PICK THE BEST OF THE TWO OPTIONS, WHICH WOULD DEPEND ON THE CASE AS THESE 2 DATASETS SHOW. THIS COULD GO UNDER FUTURE WORK, BUT IT IS BETTER TO ALREADY HAVE A FIRST SUGGESTION ON HOW THIS MAY BE DONE. MAYBE IT COULD BE LEARNED WHAT TO DO... I HAVE TO GO NOW, NEXT PAPER WAITING} 

\section{Conclusion and Future Work}
\label{sec:con}
We have presented a hierarchical regression scheme conditioned on local reference frames. We utilize the local surface normal both as a feature map for regression and to establish the local reference frame. We also proposed an efficient surface normal estimation method based on random forests. Our system shows excellent results on two real-world, challenging datasets and is either comparable or outperforms state-of-the-art methods in hand pose estimation.

The surface normal serves as an important local property of the point cloud. While random forests are an efficient way of estimating the normal, they are only one way and other methods could be developed to be more accurate.  Given the success of using surface normals in our work, we expect that there will be benefits for model-based tracking as well. 

In our current work, we follow a tree-structured model of the hand.  Given the flexibility of our proposed conditioned regression forest, one can also perform hierarchical regressions with other underlying graphical models.  With different models, one could take into account the correlations and dependencies between fingers, especially with respect to grasping objects.  We leave this as future work in improving the current system.
% success in discriminative pose estimation 
%Future work can also be done to better exploit surface normals; \eg in model based one could extending the normal matching as an efficient energy function to model based tracking.  

\bibliographystyle{splncs}
\bibliography{egbib}
\end{document}